# Probabilistic Alternative to the Gower Distance:
# A Note on Deodata Predictors

**Cristian Alb**  CA.PUBLICUS@GMAIL.COM

**Abstract**
A probabilistic alternative to the Gower distance is proposed. The probabilistic distance enables the realization of a generic deodata predictor.

## 1  Introduction

The Gower distance is a popular measure used with mixed data, data sets with both quantitative and categorical attributes. The distance is a real number varying between zero (identical) and one (maximally dissimilar). The Gower distance is a complement to the Gower similarity coefficient [1]. Its value is one minus the Gower similarity coefficient.

In article "Collapsing the Decision Tree: The Concurrent Data Predictor" [2], a set of predictive algorithms are introduced, also referred to as "deodata" predictors. The article's focus is on data with categorical attributes. A suggestion for a continuous counterpart of the algorithms reads:

"*A continuous implementation uses attributes that represent continuous (numerical) variables. As such, the match column score cannot be a binary value anymore; a continuous number is more adequate.*
*One possibility is to normalize the attribute values of the column and estimate the distance between the two compared values as a function of the standard deviation separating them*" [2, p.12].

The Gower distance, although not a function of standard deviation, satisfies the requirement of being limited to the [0, 1] interval. The limits of the interval correspond to the discrete matching possibilities. Therefore, for continuous attributes, the Gower similarity coefficient could be used as the "match column score" described in [2].

## 2  Probabilistic Distance

The implementation of a continuous counterpart to the binary match column score can be achieved by means of a probabilistic distance. The probabilistic distance introduced herein quantifies the probability that the outcome of a random variable can occur in the interval delimited by the two compared values.

The probabilistic distance assumes that the attribute values are outcomes of a random variable. Presuming that the underlying probability distribution function of the attribute value is known, the probabilistic distance can be thought of as the area beneath the probability distribution graph in the interval delimited by the two compared values as shown in Fig. 1.

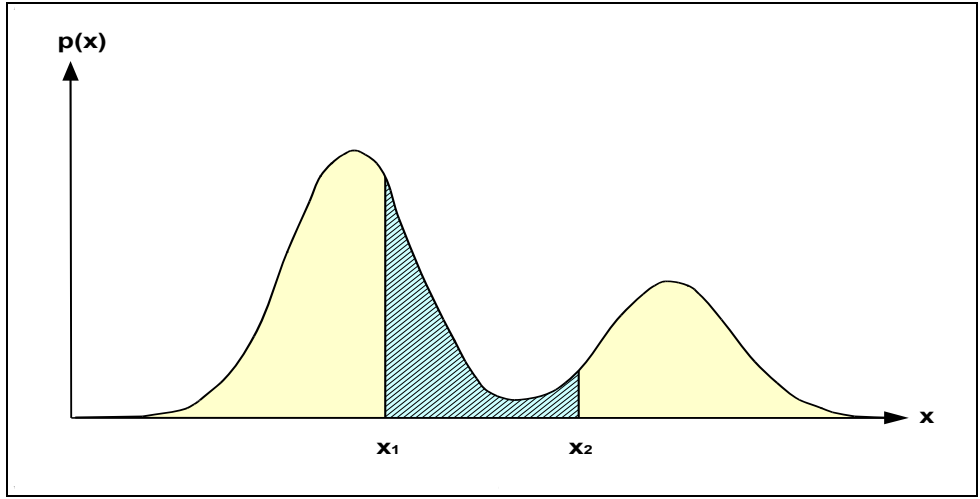

Figure 1: Area corresponding to the probabilistic distance between two values of a PDF

$$d(x_1, x_2) := P(x_1 \leq X \leq x_2) = |CDF(x_1) - CDF(x_2)| \quad (1)$$

The value of such a probabilistic distance is constrained to the interval [0,1] due to the very nature of probability distribution functions.

If the probability distribution of an attribute is not known, a normal/Gaussian distribution can be assumed. The mean and variance of the normal distribution can be estimated from the training data set. The values of the attribute can be scaled to a z-score. The Gauss error function can then be used to compute the probabilistic distance. It consists in half of the absolute difference between the outputs of the error function that correspond to the two compared values.

$$d(x_1, x_2) := \Phi(\frac{x_2 - \mu}{\sigma}) - \Phi(\frac{x_1 - \mu}{\sigma}) = \frac{1}{2} \left| erf(\frac{x_1 - \mu}{\sigma}) - erf(\frac{x_2 - \mu}{\sigma}) \right| \quad (2)$$

Being based on the concept of probability, the distance should provide a better indication of the degree of closeness between two compared values. If outlier values can be a concern for the Gower distance, the probabilistic distance should be immune to such issues by design.

The original Gower distance has been adapted to process ordinal attributes [3]. The treatment of ordinal attributes is straightforward with the probabilistic distance: the ordinal outcomes can be arranged into bins. After scaling, the bin representation approximates a probability mass function as shown in Fig. 2. The probabilistic distance is derived again as the probability of outcomes in the interval corresponding to the two compared outcomes.

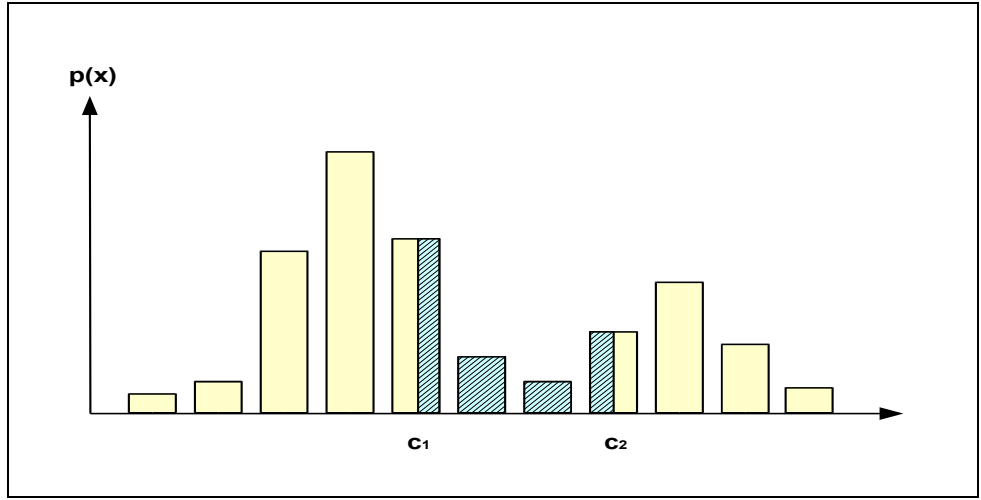

Figure 2: Area corresponding to the probabilistic distance between two ordinal outcomes

Other variations can be derived from this concept. A simple modification consists in raising the above probability distance to a power. The weight of the distance can be increased or decreased by varying the power to above or below unity:

$$d_A(x_1, x_2) := d^2(x_1, x_2) \qquad (3)$$

or:

$$d_B(x_1, x_2) := \sqrt{d(x_1, x_2)} \qquad (4)$$

The probabilistic distance was conceived as a solution to a classification problem. However, as the Gower distance, it can be used for other applications, for instance, clustering.